\pdfoutput=1

\documentclass[11pt]{article}

\usepackage[]{ACL2023}

\usepackage{times}
\usepackage{latexsym}

\usepackage{microtype}
\usepackage{graphicx}
\usepackage{subfig}
\usepackage{booktabs} 

\usepackage{placeins}
\usepackage{caption}
\usepackage{algpseudocode}
\usepackage{algorithm}
\usepackage{amsfonts}
\usepackage{amsmath}
\usepackage{latexsym}
\usepackage{graphicx}
\usepackage{multicol}
\usepackage{multirow}
\usepackage{xcolor}
\usepackage{booktabs}
\usepackage{tabularx}

\usepackage{hyperref}

\usepackage[T1]{fontenc}

\usepackage[utf8]{inputenc}

\usepackage{microtype}

\usepackage{inconsolata}

\title{STA: Self-controlled Text Augmentation for Improving Text Classifications}

\author{Congcong Wang$^\dagger$\thanks{*The author completed this work during his internship at Huawei Ireland Research Center.}, Gonzalo Fiz Pontiveros$^\ddagger$, Steven Derby$^\ddagger$ \and Tri Kurniawan Wijaya$^\ddagger$ \\
  $^\dagger$ School of Computer Science, University College Dublin, Dublin 4, Ireland \\
  $^\ddagger$ Huawei Ireland Research Centre. Georges Court, Townsend St, Dublin 2, D02 R156, Ireland\\
  \texttt{congcong.wang@ucdconnect.ie}
  \\
  \texttt{\{gonzalo.fiz.pontiveros, steven.derby, tri.kurniawan.wijaya\}@huawei.com}\\
  }

\begin{document}
\maketitle
\begin{abstract}

Despite recent advancements in Machine Learning, many tasks still involve working in low-data regimes which can make solving natural language problems difficult. Recently, a number of text augmentation techniques have emerged in the field of \emph{Natural Language Processing} (NLP) which can enrich the training data with new examples,  though they are not without their caveats. For instance, simple rule-based heuristic methods are effective, but lack variation in semantic content and syntactic structure with respect to the original text. On the other hand, more complex deep learning approaches can cause extreme shifts in the intrinsic meaning of the text and introduce unwanted noise into the training data. To more reliably control the quality of the augmented examples, we introduce a state-of-the-art approach for \emph{Self-Controlled Text Augmentation} (STA). Our approach tightly controls the generation process by introducing a self-checking procedure to ensure that generated examples retain the semantic content of the original text. Experimental results on multiple benchmarking datasets demonstrate that \emph{STA} substantially outperforms existing state-of-the-art techniques, whilst qualitative analysis reveals that the generated examples are both lexically diverse and semantically reliable.

\end{abstract}

\section{Introduction}
A variety of tasks such as \emph{Topic Classification} \citep{li-roth-2002-learning}, \emph{Emotion Detection} \citep{saravia-etal-2018-carer} and \emph{Sentiment Analysis} \citep{socher-etal-2013-recursive} have become important areas of research in NLP. Such tasks generally require a considerable amount of accurately labelled data to achieve strong performance. However, acquiring enough such data is costly and time consuming and thus rare in practice. This has motivated a vast body of research in techniques that can help alleviate issues associated with low-data regimes. 


A popular augmentation approach involves the use of rule-based transformations, which employ intuitive heuristics based on well-known paradigmatic relationships between words. For instance, by using a lexical-semantic database such as \emph{WordNet} \citep{miller1995wordnet}, researchers can make rational and domain-specific conjectures about suitable replacements for words from lists of known synonyms or hyponyms/hypernyms \citep{wang2015s, wei2019eda, feng2020genaug}. Whilst these substitution-based approaches can result in novel and lexically diverse data, they also tend to produce highly homogeneous structures, even when context-free grammars are used to generate more syntactically variable examples~\citep{jia2016data}.  

The recent success of pretrained transformer language models such as BERT \citep{devlin2019bert} and GPT-2 \citep{radford2019language} has helped facilitate more robust strategies for dealing with low-resource scenarios: Conditional text generation. Large language models --- typically trained on a vast corpus of text --- contain a rich understanding of syntactic structure and semantic phenomena in the corpus and thus can be well suited for faithful domain-specific generation. Indeed, large language models have been conditioned to great success \citep{kobayashi2018contextual, wu2019conditional, anaby2020not, kumar2020data} to synthesize highly diverse training examples and strong downstream performance . 
The trade off for diverse neurally-generated data is that semantic discrepancies can emerge which can cause samples to be misaligned with their appropriate label. Ideally, the optimal augmentation method would be one that satisfies both \textbf{Lexical/Syntactic Diversity} and \textbf{Semantic Fidelity} (reliable alignment between semantic meaning and class label). 


In this paper, we propose a novel strategy --- self-controlled text augmentation (STA) that aims to tightly control the generation process in order to produce diverse training examples which retain a high level of semantic fidelity. 
Following previous work, we fine-tune a state-of-the-art sequence-to-sequence transformer model, known as \emph{T5} \citep{raffel2020exploring}, using a dataset containing only a limited number of samples and generate new samples using task-specific prompting, which has been shown to be effective in low-resource scenarios \citep{le2021many}. While similar approaches have been deployed in previous work \cite{anaby2020not}, our novel strategy effectively utilizes \emph{Pattern-Exploiting Training} \citep{schick2021exploiting, schick2021few} by employing templates of verbalization-patterns that simultaneously direct the generation process and filter noisy labels. Experimental results on multiple benchmarks demonstrate that STA outperforms existing state-of-the-art augmentation techniques. Furthermore, examining the quality of the augmented data reveals better diversity and fidelity as compared to the existing techniques.



    
    
    



\section{Related Work}

Data augmentation for text classification has been widely developed in the literature. \citet{zhang2015character} demonstrated that replacing words or phrases with lexically similar words such as synonyms or hyponyms/hypernyms is an effective way to perform text augmentation with minimal loss of generality. The authors identify the target words according to a predefined geometric distribution and then replace words with their synonyms from a thesaurus. Similarly, \citet{wei2019eda} proposed EDA (\textit{Easy Data Augmentation}) for text classification that generates new samples from the original training data with four simple operations; synonym replacement, random insertion, random swap, and
random deletion, while \citet{feng2020genaug} further explores these substitution techniques, particularly for text generation. \citet{wang2015s} instead exploit the distributional knowledge from word embedding models to randomly replace words or phrases with other semantically similar concepts. \citet{kobayashi2018contextual} built upon this idea by replacing words based on the context of the sentence, which they achieve by sampling words from the probability distribution produced by a bi-directional LSTM-RNN language model at different word positions. 

Back translation is another method that has shown to be effective for augmentation, particularly for transforming the structure of the text ~\cite{sennrich2016improving,shleifer2019low}. Here, novel samples are generated by translating the original sentence to a predetermined language, before it is eventually translated back to the original target language. More recently, researchers have looked to capitalize on the success of pretrained transformer-based language models by performing conditional text augmentation to generate new sentences from the original training data. For example, \cite{wu2019conditional} leveraged the masked language model of BERT conditioned on labeled prompts that are prepended to the text. \citet{anaby2020not} was also successfully able to finetune GPT-2 with scarcely labeled training data to generate novel sentences of text, which improved performance on downstream classification tasks. Furthermore, the authors aimed to directly tackle the label misalignment problem by filtering noisy generated sentences using a jointly trained classifier, with some success. Similar work was performed by \citet{kumar2020data} who studied conditional text augmentation using a broader range of transformer-based pre-trained language models including autoregressive models
(GPT-2), auto-encoder models (BERT), and seq2seq models (BART), the latter of which outperformed other data augmentation methods in a low-resource setting. 

Recently,  \citet{wang2021towards} also proposed using GPT-3 for text augmentation with zero-label learning, with results that were competitive when compared to fully supervised approaches. More closely related to our instruction-based generation strategy, \citet{schick2021few} propose GenPet which is used to directly tackle a number of text generation tasks rather than text augmentation itself. In their work, which builds upon previous research PET \citep{schick2021exploiting}, the authors alter the text inputs to form cloze-style questions known as prompting training~\cite{liu2021pre}, demonstrating improved performance on few-shot downstream tasks. More recent and closely aligned with our work includes both LM-BFF~\cite{gao2021making} and DART~\cite{zhang2022differentiable}.

Unlike previous work, our novel approach can successfully generate diverse samples using task-specific templates --- verbal prompts for generation and classification which signal the models objective. To ensure semantic fidelity, the model itself (self-controlling) is then used to both generate novel data and selectively retain only the most convincing examples using a classification template.

\section{Method}

\begin{figure*}[h!]
    \centering
    \includegraphics[scale=0.8]{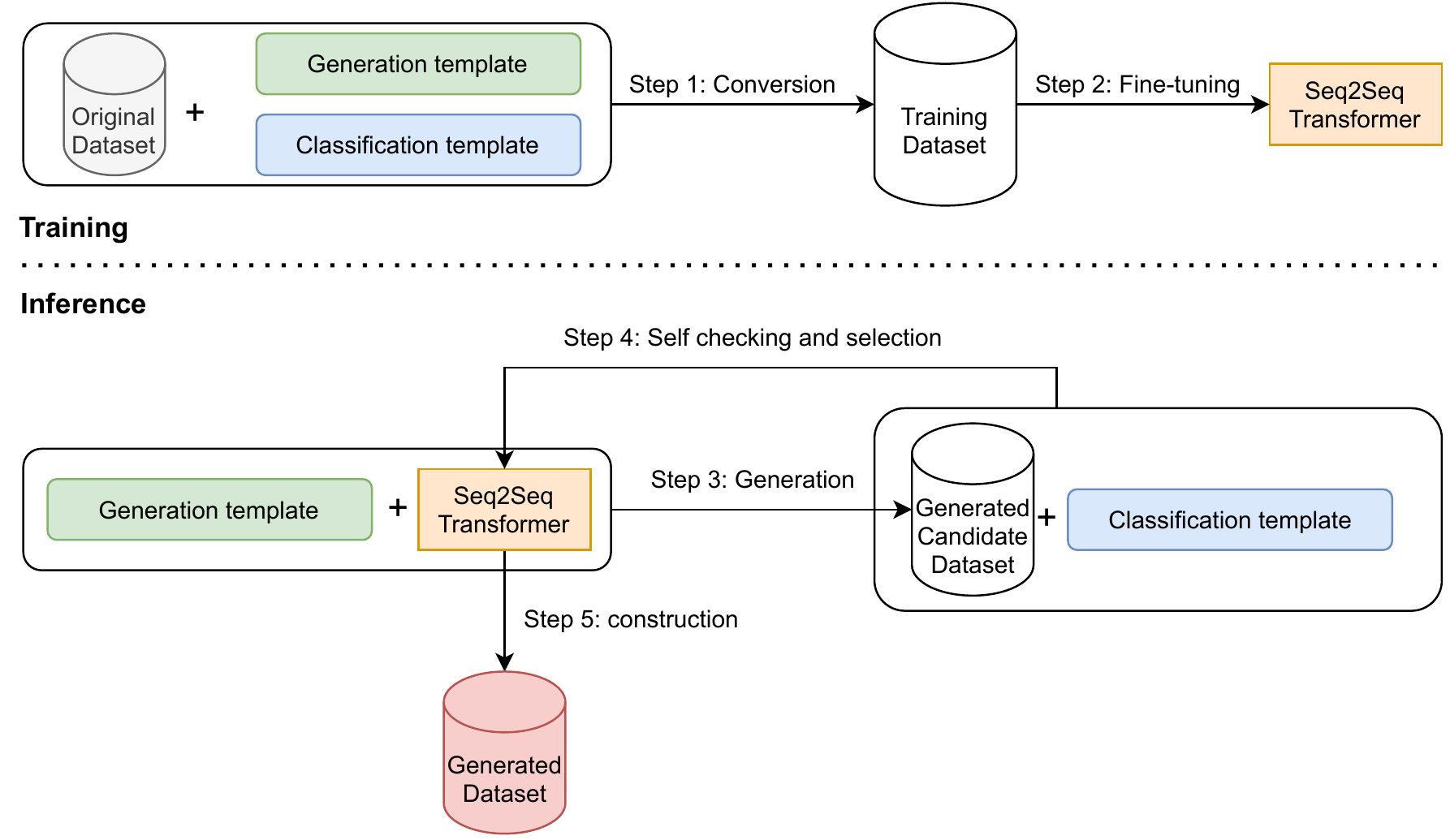}
    \caption{The architecture of our Self-controlled Text Augmentation approach (STA). The upper portion outlines the finetuning component of our method (\textbf{Training}), whilst the lower portion demonstrates our procedure for generating novel data (\textbf{Inference}). STA is highlighted by using the generation template and classification template for fine-tuning a seq2seq transformer model. The generation template is used for generating samples and the classification template is used for self-controlling and selecting the generated samples.}
    \label{fig:ste-method}
\end{figure*}
In this section, we describe our self-controlled approach for text augmentation in text classification (STA). Figure~\ref{fig:ste-method} illustrates the workflow of STA and Algorithm~\ref{alg:sta-alg} states STA in simple terms. At a high level, STA first finetunes a pretrained sequence-to-sequence (seq2seq) model using a dataset which implicitly includes generation and classification tasks. The generation task is then employed to generate new data, and the classification task is used for self-checking and selection for the final synthetic dataset.

\begin{algorithm}
\caption{:Self-Controlled Text Augmentation}\label{alg:sta-alg}
\begin{algorithmic}[1]

\Require Original dataset $\mathcal{D}_o$. Generative model $M$. Generation template $\mathcal{G}$. Classification template $\mathcal{C}$.

\State Convert $\mathcal{D}_o$ to training dataset $\mathcal{D}_t$ via $\mathcal{G}$ and $\mathcal{C}$.
\State Finetune $M$ on $\mathcal{D}_t$ in a generation task and a classification task jointly to obtain $M_t$.
\State Use $\mathcal{G}$ and $M_t$ to generate candidate dataset $\mathcal{D}_c$.
\State Apply $M_t$ to do classification inference on $\mathcal{D}_c$ with $\mathcal{C}$ to select the most confident examples.
\State Form the final generated dataset $\mathcal{D}^*$ with the selected examples.

\end{algorithmic}
\end{algorithm}
\begin{table*}[]
\centering
\small
\begin{tabular}{c|ll|c}
\toprule
\multicolumn{1}{l}{}        \textbf{Template}               &        & \textbf{Source sequence ($s$)}                                                              & \textbf{Target sequence ($t$)}                 \\
\midrule
\multirow{3}{*}{Classification }       & $c_1$ & Given \{Topic\}: \{$\mathcal{L}$\}. Classify: \{$x_i$\}                                    & \{$y_i$\}                           \\
                                           & $c_2$ & Text: \{$x_i$\}. Is this text about \{$y_i$\} \{Topic\}?                         & yes                                 \\
                                           & $c_3$ & Text: \{$x_i$\}. Is this text about \{$\overline{y}_i$\} \{Topic\}?              & no                                  \\
                                           \midrule
\multirow{2}{*}{Generation} & $g_1$ & Description: \{$y_i$\} \{Topic\}. Text:                                          & \{$x_i$\}                           \\
                                           & $g_2$ & Description: \{$y_i$\} \{Topic\}. Text: \{$x_j$\}. Another text: \{$x_i^{0\text{-}2}$\} & \{$x_i^{3...}$\} \\
                                           \bottomrule

\end{tabular}
\caption{Prompt templates. ``Topic'' refers to a simple keyword describing the dataset e.g. ``Sentiment'' or ``Emotion'' and  $\mathcal{L}$ is the list of all class labels in the dataset. The symbol $\overline{y}_i$ in $c_{3}$ stands for any label in $\mathcal{L}\setminus\{y_i\}$, chosen randomly. In $g_{2}$, the $x_j$ denotes another sample from  the same class as $x_i$ (i.e. $y_j=y_i$) chosen randomly.}
\label{tab:templates}

\end{table*}    



\subsection{Pattern-Exploiting Training in seq2seq Models}

Pattern-Exploiting Training, PET \cite{schick2021exploiting}, is a finetuning technique for downstream text classification tasks in masked language models. The authors in \citep{schick2021exploiting} show PET allows accurate text classification with very few labeled examples by converting inputs into cloze questions. In this paper we adapt the principles of PET to seq2seq autoregressive models.

Let $M$ be  a pretrained seq2seq autoregressive transformer model (for our experiments we have chosen  T5~\cite{raffel2020exploring}). Such models consist of an encoder-decoder pair; the encoder takes an input sequence $s$ and produces a contextualised encoding sequence $ \overline{s}$. The encoded sequence and current subsequence $t\colon\{t_1,t_2,..t_{i-1}\}$ are then used as the input for the decoder to compute the conditional distribution  $p_{M}(t_{i} | t_{1: i-1},\overline{s})$ for the next token in the sequence. It is the possible target sample (a sequence) $t\colon\{t_1,t_2,...,t_m\}$ given $\overline{s}$ via the factorization:


\begin{equation}
\label{eq:conditional-generate}
    p_{M}(t_{1:m} | \overline{s}) = \prod_{i=1}^{m} p_{M}(t_i | t_{1: i-1},\overline{s})
\end{equation}

Let $\mathcal{D}_o= \{(x_i,y_i)\}|_{i=1}^n$ be a dataset for text classification where $x_i \in \mathcal{X}$ and $y_i \in \mathcal{L}$ are text and label respectively. The goal is to produce a derived dataset $\mathcal{D}_t$ to finetune $M$ and ensure it is primed for generating diverse and (label) faithful examples.

Formally, a \emph{template} is a function $T: V^* \times \mathcal{L} \to V^* \times V^*$ where $V$ is the vocabulary of $M$ and $V^*$ denotes the set of finite sequences of symbols in $V$. Given a family of templates $\mathcal{T}$, we set $\mathcal{D}_t = \mathcal{T}(\mathcal{D}_o) = \bigcup_{T\in \mathcal{T}}T(\mathcal{D}_o)$. That is, we convert each sample $(x_i,y_i) \in \mathcal{D}_o$ to $|\mathcal{T}|$ samples in the derived dataset $D_t$. Table~\ref{tab:templates} lists all the templates we specifically designed for classification and generation purposes~\footnote{We have a discussion on why we use this specific set of prompts in Section~\ref{limitations}.} and Table~\ref{tab:demo-ex} (see Appendix~\ref{appendix:template}) demonstrates how this conversion is performed. 


Crucially, we construct two types of template families: classification templates $\mathcal{C}$ and generation templates $\mathcal{G}$ and set $\mathcal{T} = \mathcal{C}\cup\mathcal{G}$.

\textbf{Classification templates} have the form $c(x,y) = (f_1(x),f_2(y))$ or $c(x,y) = (f_1(x,y),f_2(y))$ where $f_1$ and $f_2$ denote functions that map a piece of text to a source sequence and target sequence respectively. Here, the text $x \in \mathcal{X}$ is not a part of the target output. \textbf{Generation templates} have the form $g(x,y) = (f_1(y),f_2(x))$ or $g(x,y) = (f_1(x,y),f_2(x))$ i.e. the label $y \in \mathcal{L}$ is not a part of the target output. Thus $D_t$ is designed so that our model can learn both how to generate a new piece of text of the domain conditioning on the label description as well as to classify a piece of text of the domain. With the dataset $D_t$ in hand, we proceed to finetune $M$ to obtain $M_t$, see \ref{sec:train_eval_details} for details on training parameters. We next describe how to use $M_t$ for text generation and self-checking.


\subsection{Data Generation, Self-checking and Selection}
\label{subsec:self-check}
We follow a two-step process: first we generate candidates and second we select a fraction of the candidates to be included as augmentations. This processes is conducted for each class separately so we may assume for the remainder of this section that we have fixed a label $y \in \mathcal{L}$. 

That is, first, we generate $\alpha\times n_y$ samples  where $n_y$ is the original number of samples in $\mathcal{D}_o$ for label $y$ and then select 
select the top $\beta \times n_y$ samples ($\beta < \alpha$). In our experiments, we call $\beta$ the \emph{augmentation factor} and  set $\alpha = 5 \times \beta$. Namely, our self-checking technique selects the top $20\%$ of the candidate examples per class~\footnote{This is based on our experimental search over \{10\%, 20\%, 30\%, 40\%, 50\%\}.} to form the final generated $D^*$ that is combined with the original dataset $D_o$ for downstream model training.

For the generation task, we need to choose a prefix/source sequence $s$ and proceed autoregressively using Equation~\ref{eq:conditional-generate}.

Referring back to Table~\ref{tab:templates}, there are two choices $g_{1}$ and $g_{2}$ that can be used to construct $s$. We choose $g_{1}$ over $g_{2}$ as the former only needs the label (the dataset description is viewed as a constant), i.e.  
$$g_1(x,y) = (f_1(y),f_2(x)).$$ 
which gives the model greater freedom to generate diverse examples. 

Thus we set $s=f_{1}(y)$ and generate $\alpha\times n_y$ samples using the finetuned model $M_t$ where $\alpha$ is the times of the number of generated candidate examples to that of original examples.

We now possess a synthetic candidate dataset $\mathcal{D}_c^y = \{(x_i,y)\}|_{i=1}^{\alpha \times n_y}$ which we will refine using a self-checking strategy for selecting the generated samples based on the confidence estimated by the model $M_t$ itself. 

For each synthetic sample $(x,y)$, we construct a source sequence using the template $c_1(x,y)= (f'_1(x),f'_2(y))=(f'_1(x),\{y\})$, that is, we set $s=f'_{1}(x)$. Given $s$  we define a score function $u$ in  the same way as  \citep{schick2021exploiting}:
$$u(y|s) =  \log p_{M_t}(\{y\}|\overline{s})$$
equivalently this is the \emph{logit} computed by $M_t$ for the sequence $\{y\}$. We then renormalize over the labels in $\mathcal{L}$ by applying a softmax over each of the scores $u(\cdot|s)$:

$$q(y|s) = \frac{e^{u(y|s)}}{\sum_{l\in \mathcal{L}}e^{u(l|s)}}$$

Finally, we rank the elements of  $\mathcal{D}_c^y$ by the value of  $q$ and select the top $\beta \times n_y$ samples to form the dataset $D_*^{y}$ and set $D_* = \bigcup_{y \in \mathcal{L}}D_*^{y}$



\begin{table*}[h!]
\centering
\small
\begin{tabular}{l|lllll}
\toprule
    \textbf{Augmentation Method}                          &  \textbf{5}                    & \textbf{10}                   & \textbf{20}                   & \textbf{50}                   & \textbf{100}                  \\
    
                              \midrule \specialrule{.1em}{.05em}{.05em} 
                              Baseline (No   Aug.) & 56.5 \, (3.8)           & 63.1 \, (4.1)           & 68.7 \, (5.1)          & 81.9 \, (2.9)          & 85.8 \, (0.8)          \\
                             \midrule 
EDA~\cite{wei2019eda}                  & 59.7\, (4.1)          & 66.6\, (4.7)          & 73.7\, (5.6)          & 83.2\, (1.5)          & 86.0\, (1.4)           \\
BT~\cite{edunov2018understanding}                   & 59.6\, (4.2)          & 67.9\, (5.3)          & 73.7\, (5.8)          & 82.9\, (1.9)          & 86.0\, (1.2)          \\
BT-Hops~\cite{shleifer2019low}              & 59.1\, (4.6)          & 67.1\, (5.2)           & 73.4\, (5.2)          & 82.4\, (2.0)          & 85.8\, (1.1)          \\
CBERT~\cite{wu2019conditional}                & 59.8\, (3.7)            & 66.3\, (6.8)          & 72.9\, (4.9)          & 82.5\, (2.5)          & 85.6\, (1.2)          \\

GPT-2~\cite{kumar2020data}                 & 53.9\, (2.8)          & 62.5\, (3.8)           & 69.4\, (4.6)          & 82.4\, (1.7)           & 85.0\, (1.7)          \\
GPT-2-$\lambda$~\cite{anaby2020not}          & 55.4\, (4.8)& 	65.9\, (4.3)& 	76.2\, (5.6)& 	84.5\, (1.4)	& 86.4\, (0.6)
          \\
BART-Span~\cite{kumar2020data}            & 60.0\, (3.7)          & 69.0\, (4.7)          & 78.4\, (5.0)          & 83.8\, (2.0)           & 85.8\, (1.0)          \\
\midrule
STA-noself   & 66.7\, (5.0)   &	77.1\, (4.7)	  & 81.8\, (2.1)   &	84.8\, (1.0)   &	85.7\, (1.0)

\\
STA-twoprompts   & 69.8\, (4.9) &	 79.1\, (3.4)	 & 81.7\, (4.5) &	\textbf{86.0\, (0.8)} &	\textbf{87.5\, (0.6)}
\\

\textbf{STA (ours)}       & \textbf{72.8\, (6.2)} & \textbf{81.4\, (2.6)} & \textbf{84.2\, (1.8)} & \textbf{86.0\, (0.8)} & 87.2\, (0.6) \\
\bottomrule
\end{tabular}
\caption{STA on \textbf{SST-2} in $5, 10, 20, 50, 100$ examples per class. The results are reported as average (std.) accuracy (in \%) based on $10$ random experimental runs. Numbers in \textbf{bold} indicate the highest in columns.}
\label{sst2-res}
\end{table*}

\section{Experiments}

Next, we conduct extensive experiments to test the effectiveness of our approach in low-data regimes. This section first describes the datasets choices, and then presents the baselines for comparison, and finally outlines model training and evaluation.



\subsection{Datasets}

Following previous work in the augmentation literature~\cite{kumar2020data,anaby2020not}, two bench-marking datasets are used in our experiments: \textbf{SST-2}~\cite{socher-etal-2013-recursive} and \textbf{TREC}~\cite{li-roth-2002-learning}. We also include \textbf{EMOTION} (emotion classification)~\cite{saravia-etal-2018-carer} and \textbf{HumAID} (crisis tweets categorisation)~\cite{alam2021humaid}
to extend the domains of testing STA's effectiveness. More information on the datasets can be found in Appendix~\ref{appendix:datasets}.



\subsection{Baselines}
We evaluate our novel strategy against a set of state-of-the-art techniques found within the literature. These approaches include a variety of augmentation procedures from rule-based heuristics to deep neural text generation. We compare STA to the augmentation techniques as they are directly related to our method in generating samples that can be used in our subsequent study for examining the quality of generated examples. We realise that our work is also related to few-shot learning approaches such as PET and LM-BFF that use few examples for text classification, we report the results of STA compared to them in Appendix~\ref{appendix:fsl-baselines}.

\textbf{Baseline (No Aug.)} uses the original training data as the downstream model training data. Namely, no augmentation is applied anywhere.

\textbf{EDA}~\cite{wei2019eda} refers to easy data augmentation that transforms an existing example by applying local word-level changes such as synonym replacement, random insertion, etc.

\textbf{BT} and \textbf{BT-Hops}~\cite{edunov2018understanding,shleifer2019low} refers to back-translation techniques. The former is simply one step back translation from English to another language that is randomly sampled from the $12$ Romance languages provided by the ``opus-mt-en-ROMANCE'' model\footnote{\url{https://huggingface.co/Helsinki-NLP/opus-mt-en-ROMANCE}} from the transformers library~\cite{wolf2019huggingface}. The latter adds random $1$ to $3$ extra languages in the back translation using the same model.

\textbf{GPT-2} \footnote{Licensing: Modified MIT License} is a deep learning method using GPT-2 for augmentation. Following~\cite{kumar2020data}, we finetune a GPT-2 base model on the original training data and then use it to generate new examples conditioning on both the label description and the first three words of an existing example.

\textbf{GPT-2-$\lambda$} is similar to GPT-2 with the addition of the LAMBDA technique from~\citet{anaby2020not}. LAMBDA first finetunes the downstream classification model on the original training data and then use it to select the generated examples by GPT-2.

\textbf{CBERT}~\cite{wu2019conditional} is a strong word-replacement based method for text augmentation. It relies on the masked language model of BERT to obtain new examples by replacing words of the original examples conditioning on the labels.

\textbf{BART-Span}~\cite{kumar2020data} \footnote{Licensing: Attribution-NonCommercial 4.0 International} uses the Seq2Seq BART model for text augmentation. Previously, it was found to be a competitive technique for data augmentation using BERT for classification (the sort of large-scale language models finetuning for classification) in low-data regimes. It is implemented as described in~\citet{kumar2020data} that finetunes the BART large model conditioning on the label names and the texts of 40\% consecutive masked words.

\subsection{Training and Evaluation}
\label{sec:train_eval_details}

When finetuning the generation model, we select the pre-trained T5 base checkpoint as the starting weights. For the downstream classification task, we finetune ``bert-base-uncased''\footnote{\url{https://huggingface.co/bert-base-uncased}} on the original training data either with or without the augmented samples. Regarding the pre-trained models, we use the publicly-released version from the HuggingFace's transformers library~\cite{wolf2019huggingface}. For the augmentation factor (i.e., $\beta$ in Section~\ref{subsec:self-check}), the augmentation techniques including ours and the baselines are applied to augment $1$ to $5$ times of original training data. In the experiments, it is regarded as a hyper-parameter to be determined. Since our work focuses on text augmentation for classification in low-data settings, we sampled 5, 10, 20, 50 and 100 examples per class for each training dataset as per~\citet{anaby2020not}. To alleviate randomness, we run all experiments $10$ times so the average accuracy along with its standard deviation (std.) is reported on the full test set in the evaluation. More information on training and evaluation refers to Appendix~\ref{appendix:training-details}.

\section{Results and Discussion}

\begin{table*}[h!]
\centering
\small
\begin{tabular}{l|lllll}
\toprule
    \textbf{Augmentation Method}                          & \textbf{5}                    & \textbf{10}                   & \textbf{20}                   & \textbf{50}                   & \textbf{100}                  \\
\midrule \specialrule{.1em}{.05em}{.05em}
Baseline (No   Aug.) & 26.7 \, (8.5)          & 28.5 \, (6.3)          & 32.4 \, (3.9)          & 59.0 \, (2.6)          & 74.7 \, (1.7)           \\
\midrule
EDA                  & 30.1 \, (6.2)          & 33.1 \, (4.3)          & 47.5 \, (5.0)          & 66.7 \, (2.7)          & 77.4 \, (1.8)          \\
BT                   & 32.0 \, (3.0)          & 37.4 \, (3.0)          & 48.5 \, (5.1)          & 65.5 \, (2.0)           & 75.6 \, (1.6)          \\
BT-Hops              & 31.3 \, (2.6)          & 37.1 \, (4.6)          & 49.1 \, (3.5)          & 65.0 \, (2.3)          & 75.0 \, (1.5)          \\
CBERT                & 29.2 \, (6.5)          & 32.6 \, (3.9)          & 44.1 \, (5.2)          & 62.1 \, (2.0)          & 75.5 \, (2.2)          \\
GPT-2                 & 28.4 \, (8.5)          & 31.3 \, (3.5)          & 39.0 \, (4.1)          & 57.1 \, (3.1)          & 69.9 \, (1.3)           \\
GPT-2-$\lambda$          & 28.6 \, (5.1)          & 30.8 \, (3.1)          & 43.3 \, (7.5)          & 71.6 \, (1.5)           & 80.7 \, (0.4)           \\
BART-Span             & 29.9 \, (4.5)           & 35.4 \, (5.7)          & 46.4 \, (3.9)          & 70.9 \, (1.5)          & 77.8 \, (1.0)          \\
\midrule
STA-noself & 34.0 \, (4.0)& 	41.4 \, (5.5)& 	53.3 \, (2.2)& 	65.1 \, (2.3)	& 74.0 \, (1.1)

\\
STA-twoprompts   & 41.8 \, (6.1) &	56.2 \, (3.0) &\textbf{64.9 \, (3.3)} &	75.1 \, (1.5)	 &81.3 \, (0.7)
\\

\textbf{STA (ours)}           & \textbf{43.8 \, (6.9)} & \textbf{57.8 \, (3.7)} & 64.1 \, (2.1)& \textbf{75.3 \, (1.8)} & \textbf{81.5 \, (1.1)} \\
\bottomrule

\end{tabular}
\caption{STA on \textbf{EMOTION} in $5, 10, 20, 50, 100$ examples per class. The results are reported as average (std.) accuracy (in \%) based on $10$ random experimental runs. Numbers in \textbf{bold} indicate the highest in columns.}
\label{emotion-res}
\end{table*}

\begin{table*}[h!]
\centering
\small
\begin{tabular}{l|lllll}
\toprule
    \textbf{Augmentation Method}                          & \textbf{5}                    & \textbf{10}                   & \textbf{20}                   & \textbf{50}                   & \textbf{100}                  \\
\midrule \specialrule{.1em}{.05em}{.05em} 
Baseline (No   Aug.) & 33.9 \, (10.4)         & 55.8 \, (6.2)          & 71.3 \, (6.3)          & 87.9 \, (3.1)          & 93.2  \, (0.7) \\
\midrule

EDA                  & 54.1 \,  (7.7)         & 70.6 \, (5.7)          & 79.5 \, (3.4)          & 89.3  \, (1.9) & 92.3 \, (1.1)          \\
BT       &           56.0 \, (8.7) &	67.0 \, (4.1)	&79.4 \, (4.8)&	89.0 \, (2.4)	& 92.7 \, (0.8)                     \\
BT-Hops      &    53.8 \, (8.2)&	67.7 \, (5.1)&	78.7  \, (5.6)&	88.0  \, (2.3)	& 91.8 \, (0.9) \\
CBERT                &    52.2 \, (9.8)	&67.0 \, (7.1)&	78.0 \, (5.3)	&89.1 \, (2.5)&	92.6 \, (1.1) \\
GPT-2                &   47.6 \, (7.9) &	67.7 \, (4.9) &	76.9 \, (5.6)	 & 87.8 \, (2.4) &	91.6 \, (1.1) \\
GPT-2-$\lambda$           &    49.6 \, (11.0)&	70.2 \, (5.8)&	80.9 \, (4.4)&	\textbf{89.6 \, (2.2)} &\textbf{93.5 \, (0.8)}           \\

BART-Span            &   55.0 \, (9.9)	&65.9 \, (6.7)&	77.1 \, (5.5)&	88.38 \, (3.4)&	92.7 \, (1.6)        \\
\midrule 
STA-noself   & 45.4 \, (3.2)  & 	61.9 \, (10.2)  & 	77.2 \, (5.5)  & 	88.3 \, (1.2)	  & 91.7 \, (0.8)
\\
STA-twoprompts   & 49.6 \, (9.0)  & 	69.1 \, (8.0)  & 	81.0 \, (5.9)  & 	89.4 \, (3.0)  & 	93.1 \, (0.9)
\\
\textbf{STA (ours)}        & \textbf{59.6 \, (7.4)}    & \textbf{70.9 \, (6.6)} & \textbf{81.1 \, (3.9)} & 89.1 \, (2.7) & 93.2 \, (0.8) \\
\bottomrule
\end{tabular}

\caption{STA on \textbf{TREC} in $5, 10, 20, 50, 100$ examples per class. The results are reported as average (std.) accuracy (in \%) based on $10$ random experimental runs. Numbers in \textbf{bold} indicate the highest in columns.}
\label{trec-res}
\end{table*}

\begin{table*}[h!]
\centering
\small
\begin{tabular}{l|lllll}
\toprule
Augmentation Method & 5                   & 10                   & 20                   & 50                   & 100                  \\\midrule \specialrule{.1em}{.05em}{.05em}
Baseline (No Aug.)  & 29.1 (6.6)         & 37.1 (6.4)          & 60.7 (4.0)           & 80.0 (0.9)          & 83.4 (1.0)          \\ \hline    
EDA                 & 49.5 (4.5)         & 64.4 (3.6)           & 74.7 (1.5)          & 80.7 (1.0)           & 83.5 (0.6)          \\
BT                   & 45.8 (5.7)         & 59.1 (5.2)          & 73.5 (2.1)          & 80.4 (1.2)          & 83.1 (0.7)           \\
BT-Hops             & 43.4 (6.4)         & 57.5 (5.2)          & 72.4 (2.8)          & 80.1 (1.1)          & 82.8 (1.4)           \\
CBERT              & 44.8 (7.6)          & 59.5 (4.8)          & 73.4 (1.7)          & 80.3 (0.8)          & 82.7 (1.2)          \\
GPT-2                & 46.0 (4.7)         & 55.7 (5.7)          & 67.3 (2.6)           & 77.8 (1.6)          & 81.1 (0.6)          \\
GPT-2-$\lambda$         & 50.7 (8.6)         & 68.1 (6.2)          & 78.5 (1.3)          & 82.1( 1.1)          & 84.2 (0.8)           \\
BART-Span           & 42.4 (7.3)          & 58.6(7.0)          & 70.04 (3.7)          & 79.3 (1.4)           & 83.33 (0.9)          \\ \hline
STA-noself          & 56.4 (7.0)         & 70.2 (4.3)          & 76.3 (3.3)          & 79.4 (4.5)          & 81.8 (1.3)          \\
STA-twoprompts      & 68.7 (10.9)        & \textbf{77.6 (3.6)} & 80.1 (1.7)          & 82.9 (1.6)           & 84.3 (0.7)          \\
\textbf{STA} (ours)          & \textbf{69.0 (3.9)} & 75.8 (3.3)          & \textbf{80.2 (1.6)} & \textbf{83.2 (0.5)} & \textbf{84.5 (1.1)} \\
\bottomrule
\end{tabular}
\caption{STA on \textbf{HumAID} in $5, 10, 20, 50, 100$ examples per class. The results are reported as average (std.) accuracy (in \%) based on $10$ random experimental runs. Numbers in \textbf{bold} indicate the highest in columns.}
\label{tab:sta-humaid-res}
\end{table*}

\subsection{Classification Tasks}
The results on \textbf{SST-2} (Table \ref{sst2-res}), \textbf{EMOTION} (Table \ref{emotion-res}), \textbf{TREC} (Table  \ref{trec-res}) and \textbf{HumAID} (Table  \ref{tab:sta-humaid-res}) classification tasks all demonstrate the effectiveness of our augmentation strategy. In all cases, our approach provides state-of-the-art performance for text augmentation across all low-resource settings. When a higher number of samples ($50\text{-}100$) are used for training we see that STA is better, as in the cases of SST-2, EMOTION and HumAID tasks, or competitive, as in the case of TREC. Furthermore, we can see that STA is superior to other augmentation techniques when only a small number of examples are used to train the generator ($5\text{-}10\text{-}20$). In fact, STA on average demonstrates a difference of $+9.4\Delta$ and $+4.7\Delta$ when trained on only $5$ and $10$ samples per class respectively, demonstrating its ability to generate salient and effective training examples from limited amounts of data.

\subsection{Ablation Studies: Self-checking and Prompts}
\label{subsec:ab-study}
To demonstrate the importance of our self-checking procedure, we performed our empirical investigations on STA both with and without the self-checking step. The results without self-checking are shown at the bottom of the tables for \textbf{SST-2} (Table \ref{sst2-res}), \textbf{EMOTION} (Table \ref{emotion-res}), \textbf{TREC} (Table  \ref{trec-res}) and \textbf{HumAID} (Table  \ref{tab:sta-humaid-res}), denoted as ``STA-noself''. We see that our approach demonstrates considerable improvements when the self-checking step is added across all tasks and training sample sizes, further supporting our augmentation technique. In fact, the difference between the two settings is considerable, with an average increase of $+9.3\Delta$ across all tasks and training samples sizes. We hypothesize that the self-checking step more reliably controls the labels of the generated text, which greatly improves training stimulus and thus the performance on downstream tasks. 

Of course, there are many possible choices for templates and permutations of template procedures. To further support the use of our multiple prompt templates used in STA (see Table~\ref{tab:templates}), we conduct another ablation run for this purpose, denoted as ``STA-twoprompts'' at the bottom of the tables. These templates, one for classification ($c_1$) and one for generation ($g_1$), represent a minimalistic approach for performing generation-based augmentation with self-checking without the additional templates outlined in Table~\ref{tab:templates}. The results show that the multiple templates used for STA provide additional improvements to the downstream tasks, especially in low-data settings.

\begin{figure*}[h!]
\centering

\subfloat[\textbf{SST-2}]{\label{fig:sst2_df}
    \includegraphics[scale=0.22]{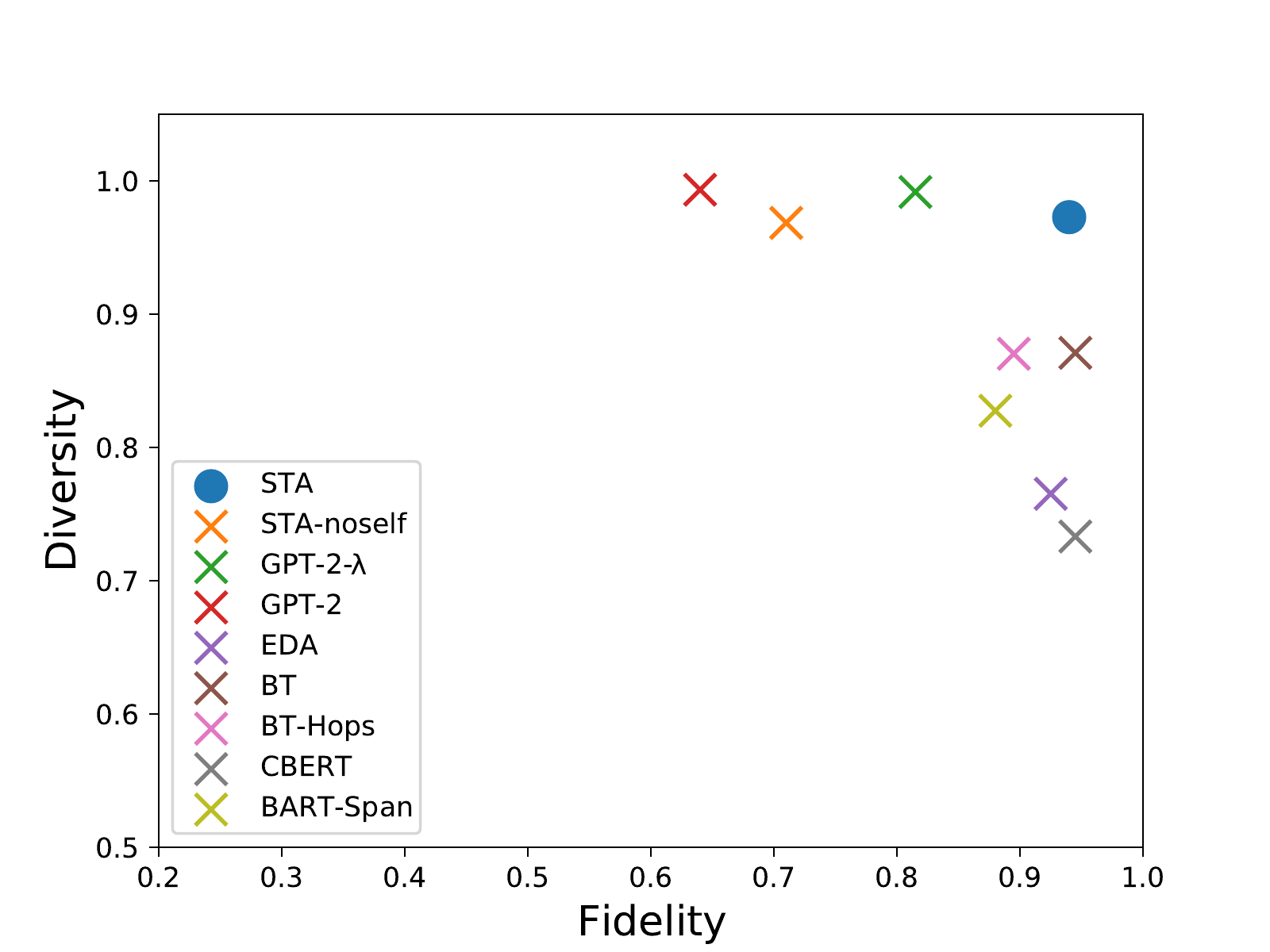}}
\subfloat[\textbf{EMOTION}]{\label{fig:emotion_df}
    \includegraphics[scale=0.22]{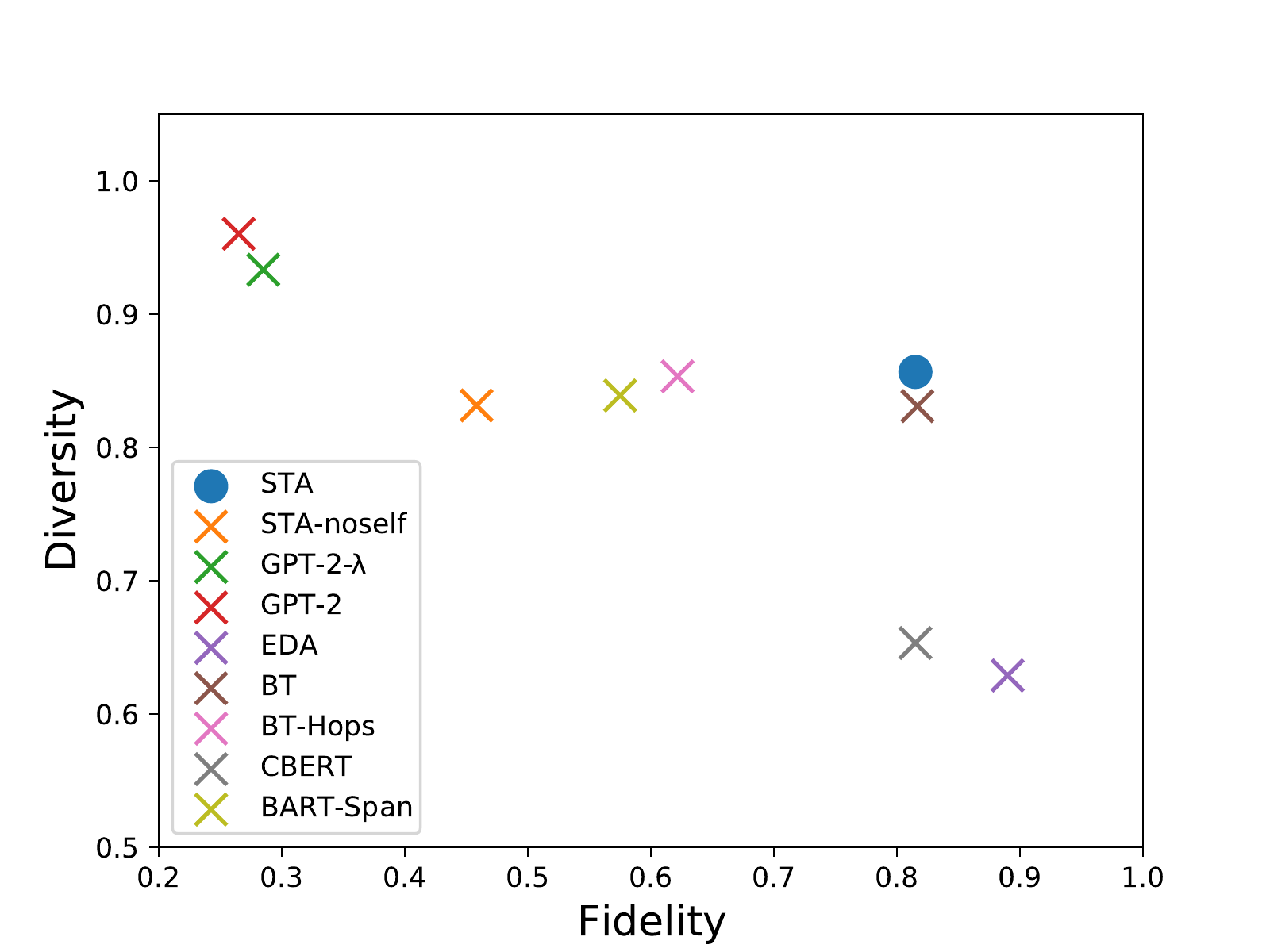}}
\subfloat[\textbf{TREC}]{\label{fig:trec_df}
\includegraphics[scale=0.22]{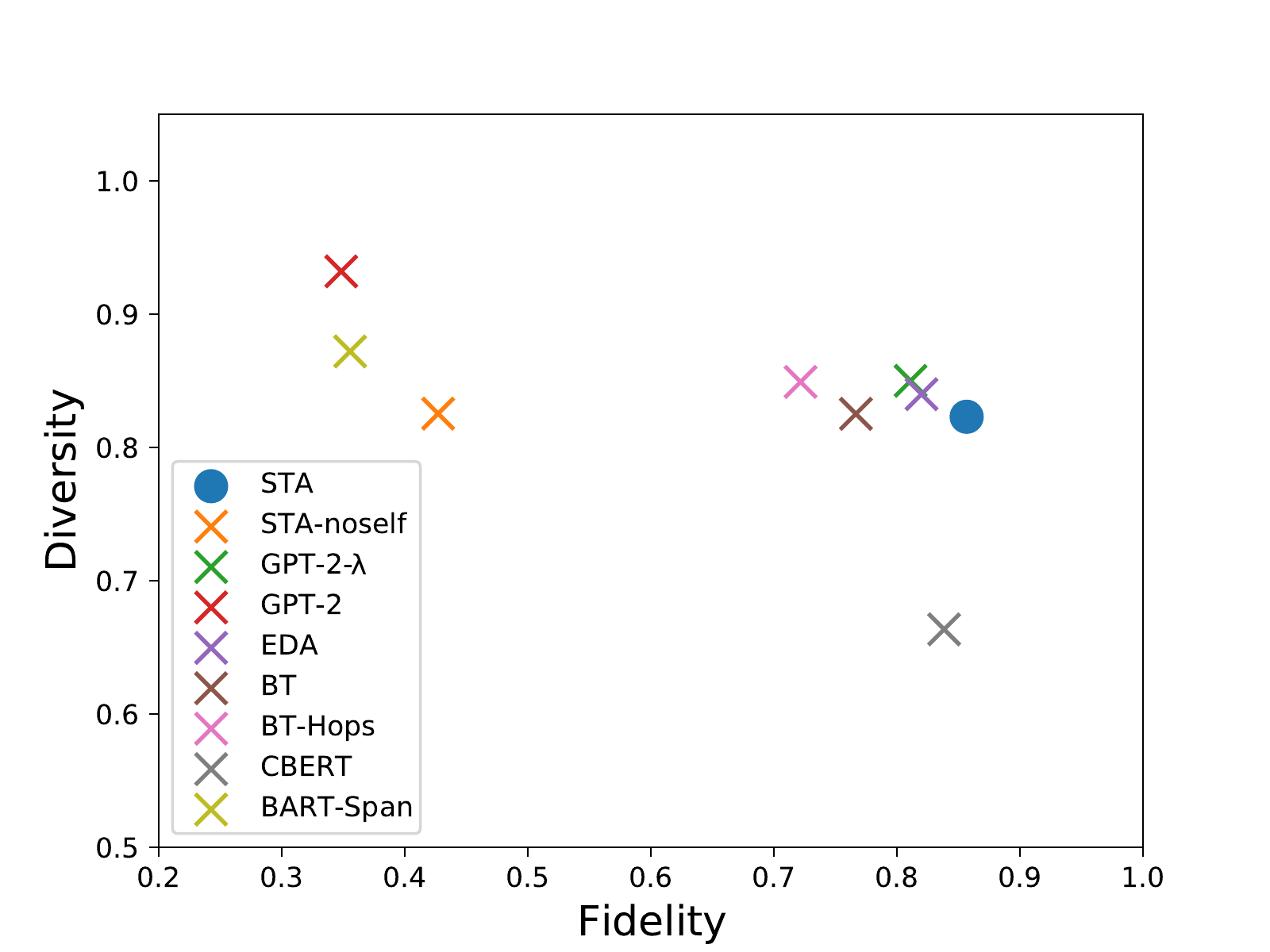}}
\subfloat[\textbf{HumAID}]{\label{fig:humaid_df}
\includegraphics[scale=0.22]{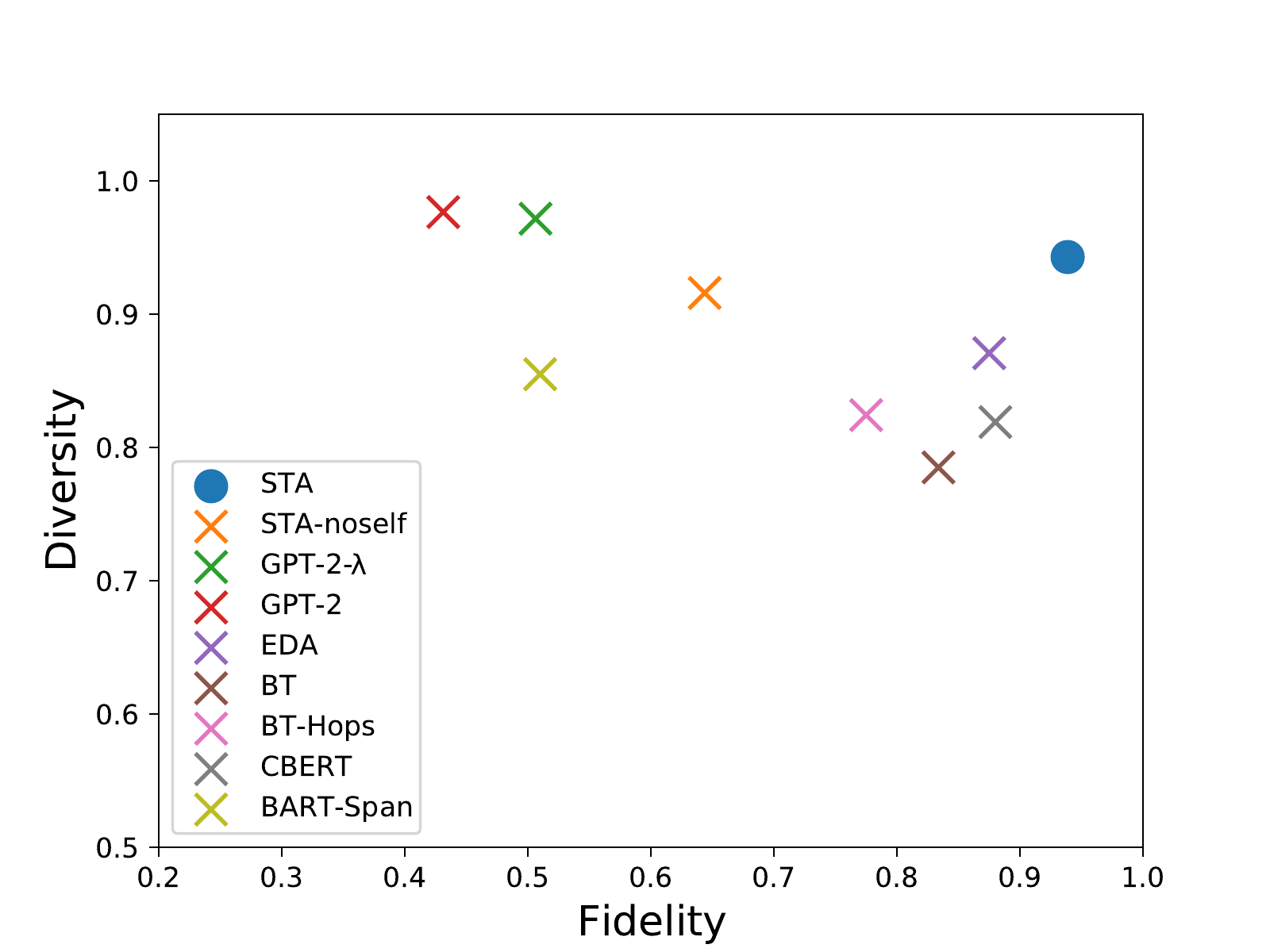}}
        \caption{Diversity versus semantic fidelity of generated texts by various augmentation methods. The average scores over 10 runs are reported.}
        \label{fig:df}
\end{figure*}

To further analyse the quality of the generated data, we measure the diversity of the data, indicated by its lexical variation, and its ability to align the text with the correct label (semantic fidelity).  The measurements for each are described as follows.
\begin{table}[h!]
    \centering
    \begin{tabular}{ccccc}
    \toprule
         & SST-2 & EMOTION & TREC & HumAID \\
       Test  & 91.8	& 93.5&	96.6 & 89.7\\
       \bottomrule
    \end{tabular}
    \caption{Accuracy (in \%) on test set predicted by BERT that is trained on the whole training data for measuring semantic fidelity.}
    \label{tab:fidelity_clas_perf}
\end{table}
\subsection{Lexical Variation and Semantic Fidelity}

\textbf{Generated Data Diversity.} The metric we used for evaluating diversity is UNIQUE TRIGRAMS~\cite{feng2020genaug,kumar2020data}. It is determined by calculating the unique tri-grams divided by the total tri-grams in a population. As we aim to examine the difference between the generated data and the original data, the population consists of both the original and generated training data. For this metric, a higher score indicates better diversity.

\textbf{Generated Data Fidelity.} The semantic fidelity is measured by evaluating how well the generated data retains the semantic meaning of its label. As per~\citet{kumar2020data}, we measure it by first finetuning a ``BERT-base-uncased '' on the 100\% of original training data of each classification task. The performance of the classifier on the test set is reported in Table~\ref{tab:fidelity_clas_perf}. After the finetuning, to measure the generated data fidelity, we use the finetuned classifier to predict the labels for the generated data and use the accuracy between its predicted labels and its associated labels as the metric for fidelity. Hence, a higher score indicates better fidelity.


To present the quality of generated data in diversity and fidelity, we take the training data (10 examples per class) along with its augmented data ($\beta=1$) for investigation. Figure~\ref{fig:df} depicts the diversity versus semantic fidelity of generated data by various augmentation methods across three datasets. We find that generation-based approaches such as GPT-2 or GPT-2-$\lambda$, achieve strong diversity but less competitive fidelity. On the contrary, rule-based heuristics methods such as EDA perform well in retaining the semantic meaning but not in lexical diversity. The merit of STA is that it is good in both diversity and fidelity, as seen from its position at the top-right of Figure~\ref{fig:sst2_df},~\ref{fig:emotion_df}, \ref{fig:trec_df} and~\ref{fig:humaid_df}. Finally, if we compare our STA approach with and without self-checking, we see that each approach produces highly diverse examples, although only self-checking STA retains a high level of semantic fidelity. As previously suggested, this ability to align the semantic content of generated examples with the correct label is the most probable reason for the increase in downstream classification performance when self-checking is employed. This supports the notion that our generation-based approach is able to produce novel data that is lexically diverse, whilst the self-checking procedure can ensure consistent label retention, which guarantees a high semantic fidelity in the generated examples\footnote{See also Appendix~\ref{appendix:demo-aug-exs} for the demonstration of augmented examples.}.

\section{Conclusion}
 We propose a novel strategy for text-based data augmentation that uses pattern-exploiting training to generate training examples and ensure better label alignment. Our approach substantially outperforms the previous state-of-the-art on a variety of downstream classification tasks and across a range of low-resource scenarios. Furthermore, we provide an analysis of the lexical diversity and label consistency of generated examples, demonstrating that our approach produces uniquely varied training examples with more consistent label alignment than previous work. In the future, we hope to improve this approach in rich-data regime and extend it to other downstream natural language tasks. TODO: add for example, this, this and this (see ACL rolling review feedback)

\section{Limitations}
\label{limitations}

Our work explores the possibility of data augmentation for boosting text classification performance when the downstream model is finetuned using pre-trained language models. The results show that STA consistently performs well across different bench-marking tasks using the same experimental setup, which addresses the limitation stated in the previous work~\cite{kumar2020data} calling for a unified data augmentation technique. However, similar to~\citet{kumar2020data}, although STA can achieve improved performance as the data size goes up to 100 examples per class in some cases (such as 100 examples per class in \textbf{EMOTION}, Table~\ref{emotion-res} and \textbf{HumAID}, Table~\ref{tab:sta-humaid-res}), the absolute gain in performance plateaus when the training data becomes richer (such as 100 examples per class in \textbf{SST-2} and \textbf{TREC}). This suggests that it is challenging for STA to improve pre-trained classifier’s model performance in more abundant data regimes. 


Another important consideration is the choice of templates used in STA. Ablation experiments in Section~\ref{subsec:ab-study} show that our chosen set of templates yields better performance than a `minimal subset' consisting of the two simplest templates; the question as to how to choose optimal templates for this augmentation scheme remains unanswered. Hence, in future work, we will explore better methods for constructing the prompt templates, aiming to reduce the dependency on the manual work at this step. 



\bibliography{anthology,custom}
\bibliographystyle{acl_natbib}

\appendix


\section{Template Example}
\label{appendix:template}

Table~\ref{tab:demo-ex} presents how an original training example is converted to multiple examples in STA using the prompt templates from Table~\ref{tab:templates}.

\begin{table*}[h!]
\small
\centering
\begin{tabular}{p{23em}|p{23em}}
\toprule \specialrule{.1em}{.05em}{.05em}

\multicolumn{2}{l}{An example from \textbf{SST-2} a sentiment  classification dataset where the classes ($\mathcal{L})$: negative, positive}                                                                                                            \\
\hline
Text ($x$)                                                                                                                           & \textit{\textcolor{blue}{top-notch action powers this romantic drama.}}

\\
 \hline  
Label ($y$)                                                                                                                       & \textit{\textcolor{red}{positive}}                                                                            \\
\midrule \specialrule{.1em}{.05em}{.05em} 
\multicolumn{2}{l}{Converted  examples by classification templates: source($s$), target($t$)}                                                                                                                                                                                                                                             \\

\hline
\textit{Given sentiment: negative, positive.   Classify: \textcolor{blue}{top-notch action powers this romantic drama.}}                                   & \textit{\textcolor{red}{positive}}                                                                            \\
\hline

\textit{Text: \textcolor{blue}{top-notch action powers this romantic drama.} Is this text about \textcolor{red}{positive} sentiment?}                                       & \textit{yes}                                                                                 \\
\hline
\textit{Text: \textcolor{blue}{top-notch action powers this romantic drama.} Is this text about negative sentiment?}                                       & \textit{no}                                                                                  \\

\midrule \specialrule{.1em}{.05em}{.05em}
\multicolumn{2}{l}{Converted  examples by generation templates: source($s$), target($t$)}                                                                                                                                                                                                              \\

\hline
\textit{Description: \textcolor{red}{positive} sentiment. Text: }                                                                                            & \textit{\textcolor{blue}{top-notch action powers this romantic drama.   }    }                             \\
\hline
\textit{Description: \textcolor{red}{positive} sentiment. Text: \textcolor{blue}{top-notch action powers this romantic drama.} Another text: spielberg 's   realization of} & \textit{a near-future america is masterful .}                                          \\
\hline
\textit{Description: \textcolor{red}{positive} sentiment. Text: \textcolor{blue}{top-notch action powers this romantic drama.} Another text: a movie in}                    & \textit{which laughter and self-exploitation merge into jolly soft-porn 'em   powerment . '} \\
\hline
\textit{Description: \textcolor{red}{positive} sentiment. Text: \textcolor{blue}{top-notch action powers this romantic drama .} Another text: a tightly   directed} & \textit{highly professional film that 's old-fashioned in all the best possible   ways . }  \\
\bottomrule
\end{tabular}
\caption{The demonstration of an example conversion by the prompt templates in Table~\ref{tab:templates} where the example's text is highlighted in \textcolor{blue}{blue} and label is highlighted in \textcolor{red}{red} for readability.}

\label{tab:demo-ex}
\end{table*}

\section{Datasets}
\label{appendix:datasets}

\begin{table}[]
\centering
\small
\begin{tabular}{l|rrr|cc}
\toprule
  \textbf{Dataset}      & \# \textbf{Train} & \# \textbf{Dev} & \# \textbf{Test} & \# \textbf{Classes ($N$)}  \\
        \midrule
SST-2    & 6,228     & 692    & 1,821    & 2    \\
EMOTION & 160,000   & 2,000   & 2,000    & 6         \\
TREC    & 4,906     & 546    & 500     & 6  \\
HumAID    & 40,623     & 5,913    & 11,508     & 8  \\
\bottomrule
\end{tabular}
\caption{Datasets statistics}
\label{tab:data-stats}
\end{table}

Table~\ref{tab:data-stats} lists the basic information of the four datasets used in our experiments and they are shortly described as follows.

\begin{itemize}
    \item \textbf{SST-2}~\cite{socher-etal-2013-recursive} is a binary sentiment classification dataset that consists of movie reviews annotated with positive and negative labels.
    \item \textbf{EMOTION}~\cite{saravia-etal-2018-carer} is a dataset for emotion classification comprising short comments from social media annotated with six emotion types, such as, sadness, joy, etc.
    \item \textbf{TREC}~\cite{li-roth-2002-learning} is a dataset for question topic classification comprising questions across six categories including human, location, etc.
    \item \textbf{HumAID}~\cite{alam2021humaid} is a dataset for crisis messages categorisation comprising tweets collected during 19 real-world disaster events, annotated by humanitarian categories including rescue volunteering or donation effort, sympathy and support, etc.
\end{itemize}

\section{Comparing to Few-shot Baselines}
\label{appendix:fsl-baselines}
Since our work explores a text augmentation approach for improving text classification in low-data regime, it is also related to few-shot learning methods that use few examples for text classification. We further conduct an experiment to compare STA to three state-of-the-art few-shot learning approaches: PET~\cite{schick2021exploiting}, LM-BFF~\cite{gao2021making}, and DART~\cite{zhang2022differentiable}. For fair comparison, we set the experiment under the $10$ examples per class scenario with $10$ random seeds ensuring the $10$ examples per class are sampled the same across the methods. Besides, we use \texttt{bert-base-uncased}\footnote{\url{https://huggingface.co/bert-base-uncased}} as the starting weights of the downstream classifier. The results are shown in Table~\ref{tab:compare-fewshot-baselines}. We found that although STA loses the best score to DART and LM-BFF on the \textbf{TREC} dataset, it substantially outperforms the few-shot baselines on \textbf{SST-2} and \textbf{EMOTION}. This tells us that STA is a competitive approach for few-shot learning text classification.

\begin{table}[]
\small
\centering
\begin{tabular}{llll}
\toprule
             & SST-2       & EMOTION    & TREC       \\
             \midrule
DART         & 66.5 (5.8) & 26.7 (3.0) & 74.0 (2.7) \\
LM-BFF       & 71.1 (9.5) & 30.2 (3.8) & \textbf{77.1 (3.0)} \\
PET          & 56.7 (0.8) & 28.4 (1.0) & 69.1 (1.1) \\
\midrule
\textbf{STA (ours)} & \textbf{81.4 (2.6)} & \textbf{57.8 (3.7)} & 70.9 (6.6)\\
\bottomrule
\end{tabular}
\caption{The comparison between STA and few-shot baselines using $10$ examples per class on \textbf{SST-2} and \textbf{EMOTION} and \textbf{TREC}. The results are reported as average (std.) accuracy (in \%) based on $10$ random experimental runs. Numbers in \textbf{bold} indicate the highest in columns.}
\label{tab:compare-fewshot-baselines}
\end{table}

\section{Training Details}
\label{appendix:training-details}
To select the downstream checkpoint and the augmentation factor, we select the run with the best performance on the development set for all methods. The hyper-parameters for finetuning the generation model and the downstream model are also setup based on the development set. Although using the full development set does not necessarily represent a real-life situation in low-data regime~\cite{schick2021exploiting,gao2021making}, we argue that it is valid in a research-oriented study. We choose to use the full development set since we aim to maximize the robustness of various methods’ best performance given small training data available. As all augmentation methods are treated the same way, we argue this is valid to showcase the performance difference between our method and the baselines. 

For all experiments presented in this work, we exclusively use \textit{Pytorch}\footnote{https://pytorch.org/} for general code and \textit{Huggingface}\footnote{https://huggingface.co/} for transformer implementations respectively, unless otherwise stated. In finetuning T5, we set the learning rate to $5 \times 10^{-5}$ using Adam~\cite{kingma2014adam} with linear scheduler ($10\%$ warmup steps), the training epochs to be $32$ and batch size to be $16$. At generation time, we use top-k ($k=40$) and top-p ($p=1.0$) sampling technique~\cite{holtzman2019curious} for next token generation. In finetuning downstream BERT, the hyper-parameters are similar to those of T5 finetuning, although the training epoch is set to be $20$. 
We set the training epochs to be as large as possible with the aim of finding the best model when trained on a small dataset, where the quality is based on performance on the development set. In our experiments, for a single run on all datasets, it takes around one day with a single Tesla P100 GPU (16GB) and thus estimated $10$ days for $10$ runs. To aid reproducibility, we will release our experimental code to the public at~\footnote{\url{https://github.com/wangcongcong123/STA}}.



\section{Demonstration}
\label{appendix:demo-aug-exs}

Table~\ref{tab:demo-aug-exs} and Table~\ref{tab:sta-humaid-demo-aug-exs} demonstrate some original examples and augmented examples by different methods. In comparison, the examples generated by STA tend to be not only diverse but also highly label relevant (semantic fidelity).


\begin{table*}[h!]
\small\centering
\begin{tabular}{c|p{43em}}
\toprule
\multicolumn{2}{l}{Original training examples and augmented examples for ``Sadness" of \textbf{EMOTION}}                                                                                                                          \\
\hline
\multirow{5}{*}{Original}        & i sit here feeling blank about this                                                                                                                                         \\

                                 & i feel ashamed that i so readily turn it   aside                                                                                                                            \\
                                 & i feel positively ashamed when i look out   of the window and see the state of things                                                                                       \\
                                 & i had just lost my uncle i would be sad   but i feel as if i am devastated                                                                                                  \\
                                 & i was feeling kind of discouraged because   nothing happened                                                                                                                \\
                                 \midrule
\multirow{5}{*}{EDA}             & i sit here opinion blank about this                                                                                                                                         \\
                                 & i feel that ashamed i so readily turn it   aside                                                                                                                            \\
                                 & i feel positively ashamed when i look out   of the window and construe the state of things                                                                                  \\
                                 & i had just lost my uncle i would be   pitiful but i feel as if i am devastated                                                                                              \\
                                 & i happened feeling kind of discouraged   because nothing was                                                                                                                \\
                                 \midrule
\multirow{5}{*}{GPT-2-$\lambda$} & ive seen so many girls walk around feeling ashamed of their bodi                                                                                                            \\
                                 & ive got to admit that i feel a little   weird for a moment seeing her standing in front of my face when i walk into   the shop                                              \\
                                 & ive always wondered what im doing right   now im feeling                                                                                                                    \\
                                 & ive read many blogs about her and how   much she hates those who don't admit to being kind or caring about others but   instead blame them for not doing something about it \\
                                 & ive never felt sympathetic towards people   because of the way they look and act because of their skin to                                                                   \\
                                 \midrule
\multirow{5}{*}{STA-noself}      & i feel like the whole world is watching and feeling it's failing me                                                                                                         \\
                                 & i want people to know i am not alone                                                                                                                                        \\
                                 & i feel ashamed when i look out of the   window and see the state of things                                                                                                  \\
                                 & i walked away feeling disappointed   because i don t know the answer                                                                                                        \\
                                 & i drank some cold drink or find some ice   dessert such as chendol or ice kacang                                                                                            \\
                                 \midrule
\multirow{5}{*}{STA}             & i feel sad seeing people who have to work harder to cope                                                                                                                    \\
                                 & i walked away feeling disappointed   because i don t know the answer                                                                                                        \\
                                 & i was feeling sad seeing the state of   things that i never did i really want to see if it lasted                                                                           \\
                                 & i feel sad seeing the state of things but   the truth is im not sure how to express it gracefully                                                                           \\
                                 & i feel like the whole world is watching   and feeling it's failing me                               \\
                                 \bottomrule
\end{tabular}

\caption{The demonstration of original training examples and augmented examples for ``sadness'' of \textbf{EMOTION}. It is noted that the $5$ augmented examples in each block are randomly selected instead of cherry-picked. This reveals some difference between the original training examples and the augmented examples by our STA and other methods (Here we use a rule-based heuristics method EDA, a generation-based method GPT-2-$\lambda$ and STA-noself for comparison).}
\label{tab:demo-aug-exs}
\end{table*}

\begin{table*}[h!]
\small
\centering
\begin{tabular}{c|p{43em}}
\toprule
\multicolumn{2}{l}{Original   training examples and augmented examples for ``missing or found people"  of HumAID}                                                                      \\
\midrule

\multirow{5}{*}{Original}     & UPDATE: Body found of man who disappeared amid Maryland flooding                                                 \\
                              & Open Missing People Search Database from   Mati and Rafina areas  \#Greecefires   \#PrayForGreece \#PrayForAthens                             \\
                              & @ThinBlueLine614 @GaetaSusan   @DineshDSouza case in point, \#California Liberalism has created the hell  which has left 1000s missing 70 dead,...    \\
                              & Heres the latest in the California wildfires \#CampFire 1011 people are missing Death toll rises to 71 Trump blames fires on poor ...   \\
                              & \#Idai victims buried in mass grave in   Sussundenga, at least 60 missing - \#Mozambique  \#CycloneIdai \#CicloneIdai                                  \\
                              \midrule
\multirow{5}{*}{EDA}          & update flooding found of man who disappeared amid maryland boy                                                                                         \\
                              & open missing people search database from   mati escape and rafina areas greecefires prayforgreece prayforathens                                        \\
                              & created gaetasusan dineshdsouza hell in   point california missing has thinblueline the case which has left s   liberalism dead an countless people... \\
                              & heres blames latest in the california   wildfires campfire people are missing death toll rises to trump more fires on poor...            \\
                              & idai victims buried in mass grave in   sussundenga at mozambique missing least cycloneidai cicloneidai                                                 \\
                              \midrule
\multirow{5}{*}{GPT-2-lambda} & @KezorNews - Search remains in \#Morocco after @deweathersamp; there has   been no confirmed death in \#Kerala                                         \\
                              & \#Cambodia - Search \& Rescue is   assisting Search \& Rescue officials in locating the missing 27 year old   woman who disappeared in ...             \\
                              & @JHodgeEagle Rescue Injured After Missing   Two Children In Fresno County                                                                              \\
                              & \#Florence \#Florence Missing On-Rescue   Teams Searching For Search and Rescue Members \#Florence \#Florence   \#DisasterInformer \#E                 \\
                              & RT @LATTAODAYOUT: RT @HannahDorian:   Search Continues After Disappearance of Missing People in Florida                                                \\
                              \midrule
\multirow{5}{*}{STA-noself}   & Search Database from Matias, Malaysia, missing after \#Maria, \#Kerala,   \#Bangladesh \#KeralaKerala, \#KeralaFloods, ...                             \\
                              & RT @hubarak: Yes, I can guarantee you   that our country is safe from flooding during the upcoming weekend!  Previous story Time Out! 2 Comments       \\
                              & The missing persons who disappeared amid   Maryland flooding are still at large. More on this in the next article.                                     \\
                              & the number of missing after \#CycloneIdai   has reached more than 1,000, reports CNN.                                                                  \\
                              & RT @adriane@przkniewskiZeitecki  1 person missing, police confirm  \#CycloneIdai. \#CicloneIdai                                                        \\
                              \midrule
\multirow{5}{*}{STA}          & The missing persons who disappeared amid Maryland flooding are still at   large. More on this in the next article.                                     \\
                              & Search Triangle County for missing and   missing after \#Maria floods \#DisasterFire                                                                   \\
                              & Just arrived at San Diego International   Airport after \#Atlantic Storm. More than 200 people were missing, including   13 helicopters ...            \\
                              & Search Database contains information on   missing and found people \#HurricaneMaria, hashtag \#Firefighter                                             \\
                              & Were told all too often that Californians  are missing in Mexico City, where a massive flood was devastating. ...             \\
                              \bottomrule
\end{tabular}
\caption{The demonstration of original training examples and augmented examples for ``missing or found people'' of \textbf{HumAID}. It is noted that the $5$ augmented examples in each block are randomly selected instead of cherry-picked. This reveals some difference between the original training examples and the augmented examples by our STA and other methods (Here we use a rule-based heuristics method EDA, a generation-based method GPT-2-$\lambda$ and STA-noself for comparison).}
\label{tab:sta-humaid-demo-aug-exs}
\end{table*}

\end{document}